\documentclass[10pt,twocolumn,letterpaper]{article}

\usepackage{cvpr}
\usepackage{times}
\usepackage{epsfig}
\usepackage{graphicx}
\usepackage{amsmath}
\usepackage{amssymb}
\usepackage{multirow}

\cvprfinalcopy 

\def\cvprPaperID{ } 

\begin{document}

\title{Diversifying Inference Path Selection: Moving-Mobile-Network for Landmark Recognition}

\author{Biao Qian$^{\dag}$
\and
Yang Wang$^{\dag}$
\and
Zhao Zhang$^{\dag}$
\and
Richang Hong$^{\dag}$
\and
Meng Wang$^{\dag}$
\and
Ling Shao$^{\S}$\\
$^{\dag}$Hefei University of Technology, China\\
 $^{\S}$Inception Institute of Artificial Intelligence, UAE\\
}

\maketitle

\begin{abstract}
Deep convolutional neural networks have largely benefited computer vision tasks. However, the high computational complexity limits their real-world applications. To this end, many methods have been proposed for efficient network learning, and applications in portable mobile devices. In this paper, we propose a novel \underline{M}oving-\underline{M}obile-\underline{Net}work, named M$^2$Net, for landmark recognition, equipped each landmark image with located geographic information. We intuitively find that M$^2$Net can essentially promote the diversity of the inference path (selected blocks subset) selection, so as to enhance the recognition accuracy. The above intuition is achieved by our proposed reward function with the input of geo-location and landmarks. We also find that the performance of other portable networks can be improved via our architecture. We construct two landmark image datasets, with each landmark associated with geographic information, over which we conduct extensive experiments to demonstrate that M$^2$Net achieves improved recognition accuracy with comparable complexity.
\end{abstract}

\section{Introduction}

Deep Convolutional Neural Networks (CNNs) have largely benefited the field of pattern recognition. However, the large complexity for training the CNNs limits their real-world applications. To this end, many works \cite{han2015learning,li2016pruning,liu2017learning,han2015deep,Cheng2016Convolutional,Cheng2016Convolutional} have been proposed on modeling compression networks, which has naturally led to a surge in research \cite{howard2017mobilenets,zhang2018shufflenet}, on applying them to real-world portable mobile devices.
For this, BlockDrop \cite{wu2018blockdrop} module is the typical model, which dynamically selects the useful inference path for efficiency.

\begin{figure}[t]
\centering
\includegraphics[width=1.0\columnwidth]{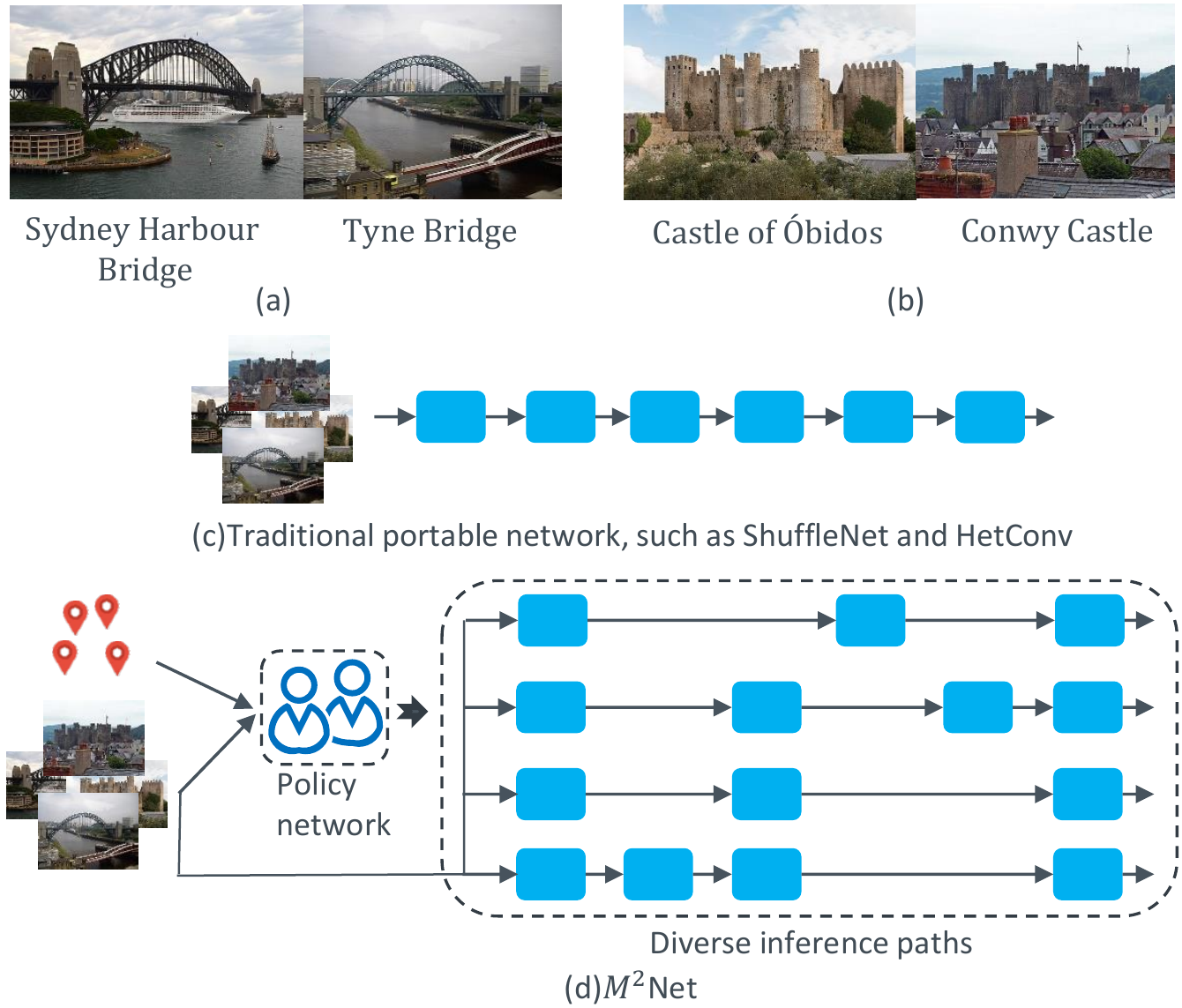}
\caption{(a)(b)Examples of similar landmarks with different geographic information, the existing  compression networks are more likely to detect them as identical landmark. (c)Traditional portable network, such as ShuffleNet \cite{zhang2018shufflenet} and HetConv \cite{Singh2019HetConv}  only generate unique inference path. (d)M$^2$Net combines geographic information with visual information to generate diverse inference paths for each input.}
\label{fig1}
\end{figure}

However, current models fail under the presence of visually similar landmarks with different geographic information, as illustrated in Fig.\ref{fig1}, where these landmarks belong to different small classes in the same large categories, \emph{e.g.,} tower from different cities. Therefore, the visual differences between them are challenging to distinguish even when using  powerful deep network representations \cite{howard2017mobilenets,zhang2018shufflenet,Singh2019HetConv,wu2018blockdrop}. In other words, these mobile nets fail to consider the mobility of mobile networks, with moving geo-locations.

\begin{figure*}[ht]
\centering
\includegraphics[width=0.88\textwidth]{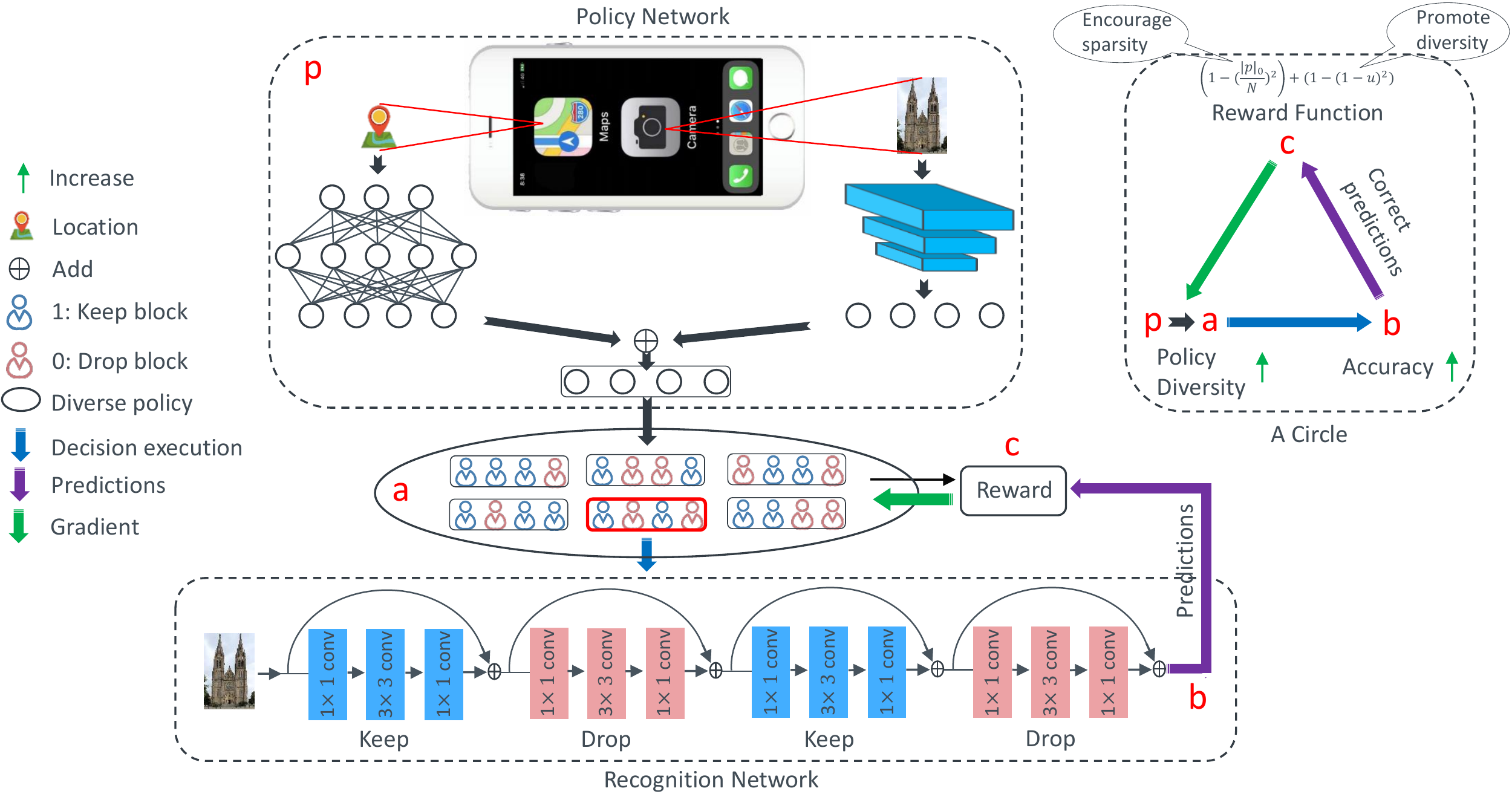}
\caption{Overview of M$^2$Net, where the blocks with light blue are kept, while red ones are dropped. For the policy network, geographic information in combination with the landmark images is exploited to generate diverse policies, to determine which blocks are kept, which dynamically promotes the diversity of the inference path selection in the recognition network. Specifically, the information from the geographic location and landmark image of mobile devices is fused. Then, an N-dimension Bernoulli distribution is utilized to perform the diverse decision to select block subsets as the inference path. The reward function encourages the selection of the inference path with minimal number of blocks and diverse output policies under correct predictions. p: Policy network. a: Output policy. b: The prediction accuracy of recognition network. c: The reward function. As discussed above, a circle module reflects promotion relationship among policy diversity(p,a), recognition accuracy(b), and reward function(c), which will be introduced in detail in the Section \ref{More_Insights}.}
\label{fig2}
\end{figure*}


To address this problem, we propose a novel moving-mobile-network, named M$^2$Net, which exploits  geographic information in combination with visual information, to improve the landmark recognition accuracy. Our basic idea is to learn a policy network based on reinforcement learning \cite{sutton2018reinforcement}, which dynamically selects the layers or blocks in the network to construct the inference path. We choose Residual Networks (ResNet) \cite{he2016deep} as our backbone networks, due to their robustness to layer removal \cite{veit2016residual}.

We remark that M$^2$Net also belongs to the BlockDrop module, yet with non-trivial extensions to tackle mobility issues of the portable network. The following observations are drawn:
\begin{enumerate}
    \item There are diverse policies,  \emph{i.e.,} more diverse inference paths can improve the accuracy of M$^2$Net.
    \item When M$^2$Net makes a correct prediction, the policy network is positively rewarded. Thus the improved accuracy will provide more rewards for the policy network.
    \item During the training process, to get the largest reward value, the policy network aims to make a sparse and unique policy, resulting in diverse inference paths for M$^2$Net.
\end{enumerate}
For ease of understanding, we illustrate the major framework of M$^2$Net in Fig.\ref{fig2}, which can promote diverse inference paths in the network, improving the recognition accuracy. Our M$^2$Net can also obtain comparable computational efficiency due to dynamic policy network output for each landmark image. The policy network is trained based on both the landmark images and geographic locations, to encourage the policy selection with less performance loss. The policy network and the recognition network are jointly learned to improve the accuracy, while achieve the feasible computational efficiency.

\begin{figure*}[t]
\centering
\includegraphics[width=0.5\textwidth]{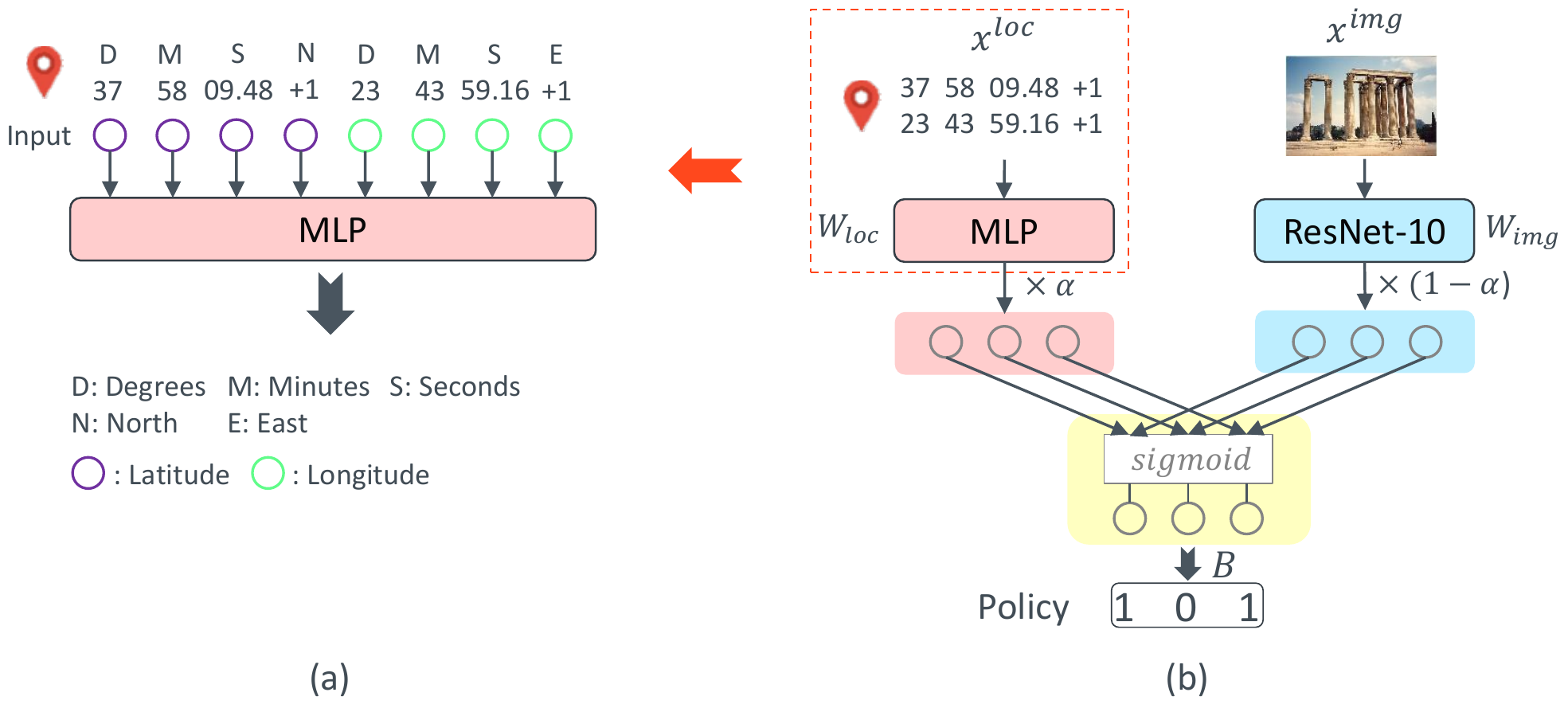}
\caption{(a)Encoding the geographic location to an 8-dimension vector is beneficial for acquiring more information from the location vector and adopting a Multi-Layer Perceptron (MLP) to extract the location features. (b)The policy network fuses geographic information and landmark images. $W_{loc}$ and $W_{img}$ are the weights of the MLP and ResNet-10, respectively. $\alpha\in[0,1]$ is the fusion factor. Sigmoid function $sigmoid(x)=\frac{1}{1+e^{-x}}$. B represents an N(N=3)-dimension Bernoulli distribution. Combining geographic information and landmark images can lead to more diverse policies.}
\label{Encoding}
\end{figure*}

Our major contributions are summarized as follows:
\begin{itemize}
\item We propose M$^2$Net to exploit the geographic information in combination with the landmark image to learn a policy network to enable the diverse inference paths selection, which improves the recognition accuracy on mobile devices. In addition, we introduce a novel reward function for the policy network to promote the generation of diverse policies.
\item For each landmark, we dynamically choose blocks, \emph{i.e.,} a subset of the inference path for the recognition network, to make the inference efficient.
\item As no existing research on compression or mobile network investigates image recognition with geo-location information, we create two such datasets named Landmark-420 and Landmark-732 for evaluations. The experimental results show that M$^2$Net achieves $4.8\%$ accuracy gain and $36\%$ speedup on average compared to the original ResNet-47 on Landmark-420, and outperforms state-of-the-art methods \cite{wu2018blockdrop,howard2017mobilenets,zhang2018shufflenet,Singh2019HetConv} in terms of the accuracy, while achieving a feasible computational complexity.
\end{itemize}

\section{Related work}
To better appreciate our research findings, we discuss related work for deep network compressions.
\subsection{Dynamic Layer Selection}
Several methods have been proposed to dynamically drop the residual layer in residual networks. Specifically, \cite{veit2016residual} found that residual networks can be regarded as a collection of short paths, and removing a single block has negligible effects on network performance. \cite{huang2018data} introduced a scaling factor to scale the residual blocks and added a sparsity constraint on these factors during the training process, removing the blocks with small factors. For SkipNet \cite{wang2018skipnet}, the gating networks are proposed to decide whether to skip the corresponding layer during inference, where the outputs of previous layers are used as the inputs of the gating networks. Unlike SkipNet that assigns a gating module for each residual layer, Blockdrop \cite{wu2018blockdrop} module characterized a policy network based on reinforcement learning to output a set of binary decisions for each block in a pre-trained ResNet, where the policy network takes images as the inputs.

Orthogonal to the above, M$^2$Net exploits both the geographic information and visual information to learn a policy network to achieve diverse policies, which can improve the performance.

\subsection{Model Compression}
Many methods aim to accelerate the network computation and reduce the model size. On one hand, some of the techniques aim to simplify the existing networks, such as network pruning \cite{han2015learning,liu2017learning,He:2018:SFP:3304889.3304970,he2019filter,chen2019layer}, quantization \cite{han2015deep,cai2017deep,wang2018two,rastegari2016xnor,xu2019main}, low-rank factorization \cite{Cheng2016Convolutional,peng2018extreme} and knowledge distillation \cite{hinton2015distilling,yim2017gift}. On the other hand, efficient network architectures are designed to train compact neural networks, such as MobileNet \cite{howard2017mobilenets}, ShuffleNet \cite{zhang2018shufflenet}, and HetConv \cite{Singh2019HetConv}.  These methods obtain compact networks after training; however, during inference, the architectures are kept unchanged for all images.

Unlike the above, M$^2$Net focuses on dynamically adjusting network architecture for different input images during inference, which contributes to allocating computing resources reasonably.

\section{M$^2$Net: Moving Mobile Network}\label{section3}

Formally, as shown in Fig.\ref{fig2},  M$^2$Net is composed of two subnetworks of the policy network, denoted as $W_{loc}$ and $W_{img}$. Given inputs of two subnetworks: the visual images $X^{img}=\{x_1^{img},x_2^{img},...,x_M^{img}\}$ and the geographic locations $X^{loc}=\{x_1^{loc},x_2^{loc},...,x_M^{loc}\}, x_i^{loc} \in \mathbb{R}^8, i \in \{1,2,...,M\}$, where M is the number of landmark samples, the policy network aims at generating binary policy vectors $P=\{p_1,p_2,...,p_M\}, p_i \in \{0,1\}^N$, where N is the number of residual blocks in recognition network.

Before shedding the light on the architecture of M$^2$Net, we first discuss how to encode the mobility of M$^2$Net, that is, geographic information, in the next section.
\begin{figure}[t]
\centering
\includegraphics[width=0.75\columnwidth]{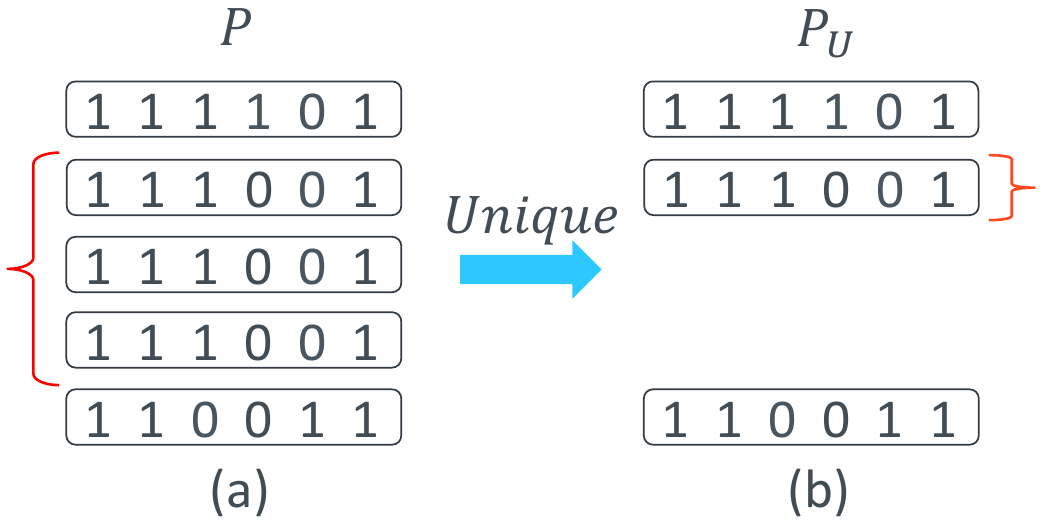}
\caption{An example of unique policy. $Unique(\cdot,...,\cdot)$ operation represents merging the same elements. $P$ is output policy of the policy network, and $P_U$ is the corresponding unique policy.  The diversity of (a) is 3.}
\label{Examples_diversity}
\end{figure}

\subsection{Encoding Geographic Location}\label{geoinfo}
Geographic location can be encoded via the following formats: Degrees Minutes Seconds ($D^{\circ}$ $M^{\prime}$ $S^{\prime\prime}$), Decimal Minutes ($D^{\circ}$ $M.M^{\prime}$), and Decimal Degrees ($D.D^{\circ}$). N, S, E or W  represent North, South, East or West, respectively.

We opt for the first format due to its higher dimension, while replacing N and S or E and W with $\pm{1}$ as flags. This can effectively extract richer information from the input. As shown in Fig.\ref{Encoding}(a), we construct an 8-dimensional vector. We observe that Degrees, Minutes and Seconds have different scales, while the sensitivity to change is of great difference. To address this, we adopt an Multi-Layer Perceptron (MLP) to extract the location features, with different layers for various scales.

With the help of geographic information, we find that M$^2$Net can intuitively promote diversity for inference path selection, which will be discussed in the next section.

\subsection{Diversity for M$^2$Net}\label{section3.1.2}
\subsubsection{Unique Policy and Diversity}
As M$^2$Net selects the inference path via the policy network, we introduce $P_U$, a set of unique policy, which is obtained by
\begin{equation}
P_U=Unique(P)=Unique(\{p_1,p_2,...,p_M\}),
\end{equation}
where $Unique(\cdot,...,\cdot)$ indicates merging the same elements in $P$, as shown in Fig.\ref{Examples_diversity}.

\textbf{Diversifying inference path selection via geo-information.} We measure the diversity as the number of unique policies,\emph{ i.e.,} the length of $P_U$, for M samples. One toy example (M=5) is shown in Fig.\ref{Examples_diversity}(a) are the policies made by the policy network. As aforementioned, the number of unique policies of (a),\emph{ i.e.,} policy diversity, is 3. As shown in Fig.\ref{Encoding}(b), given any two landmarks, denoted as $x_1^{img}$ and $x_2^{img}$, with large visual similarity, but highly distinct locations, and vice versa (similar locations yet small visual similarity), such that the outputs $W_{img}x_1^{img}$ and $W_{img}x_2^{img}$ are close, often, $x_1^{loc}$ and $x_2^{loc}$, the outputs $W_{loc}x_1^{loc}+W_{img}x_1^{img}$ and $W_{loc}x_2^{loc}+W_{img}x_2^{img}$ are usually highly different, making the fusion result \emph{i.e.,} the outputs of sigmoid function, diverse.

\begin{figure}[t]
\centering
\includegraphics[width=1.0\columnwidth]{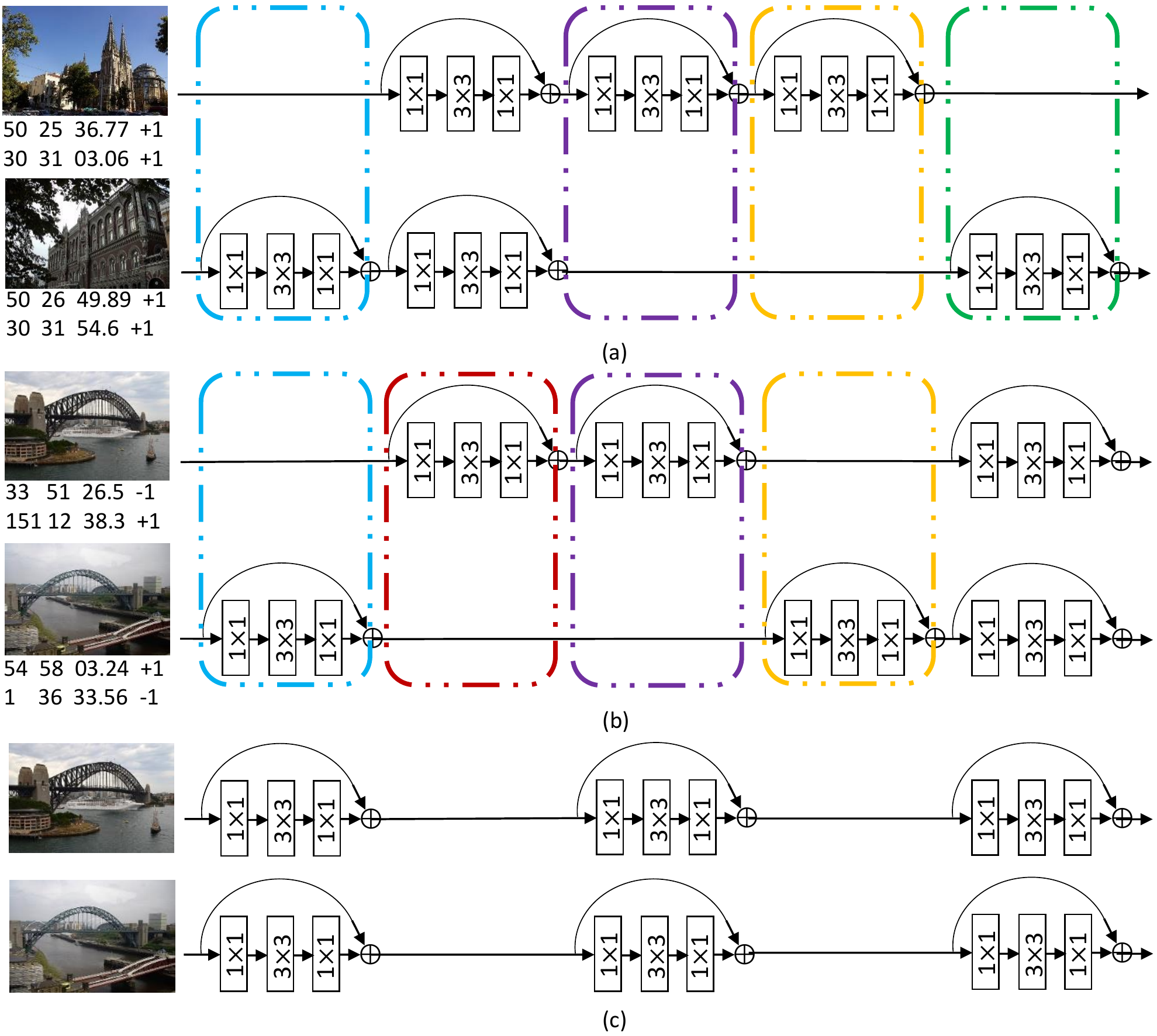}
\caption{An overview of policy diversity. (a)Similar geographic information with different landmarks. (b)Similar landmarks with different geographic information. (c)Similar landmarks without geographic information. (a)(b) and (c) come from M$^2$Net and BlockDrop, respectively. Compared with (c), our method (a)(b) makes more diverse policies (2vs1), which can improve the accuracy. In addition, (a)(b) show the difference between the inference paths, where colorful boxes mark the corresponding position, which make the inference path diverse.}
\label{overview}
\end{figure}

To exhibit the intuitions, we show some visualization results in Fig.\ref{overview} on diverse inference paths selection of M$^2$Net. The diverse policies, as shown in Fig.\ref{overview}(a) and (b), of M$^2$Net offer diverse inference paths for landmark recognition, with the help of both geographic information and landmarks, to improve the performance. Instead, Fig.\ref{overview}(c) shows that BlockDrop module can generate only one inference path.

We define the uniqueness, which describes the difference between policies of the policy network for $M$ samples, as a vector $U=\{u_1,u_2,...,u_M\}$,
\begin{equation}
\label{E_similarity}
u_i= \frac{1}{M} \sum_{j=1}^M Hamming(p_i, p_j), i \in \{1,2,...,M\},
\end{equation}
where $Hamming(\cdot,\cdot)$ denotes the normalized Hamming distance between two binary vectors. The larger $u_i$ is, the larger the difference between $p_i$ and other M-1 policies is, which indicates that $p_i$ has larger possibility to become unique policy. In addition, during the training process, we aim at promoting the diversity of output policies by increasing $U$.

To this end, we propose our reward function for the policy network, which will be discussed in the Section \ref{R_F}.

\subsection{Policy Network based on Reward Function}\label{R_F}
Motivated by \cite{wang2018skipnet} and \cite{wu2018blockdrop}, we apply reinforcement learning to train our policy network.
As shown in Fig.\ref{fig2} and Fig.\ref{Encoding}(b), given a geographic location $x^{loc}$ and landmark $x^{img}$, with a pre-trained recognition network with N residual blocks, a policy $p=\{p^1,p^2,...,p^N\}$ can be seen as an N-dimensional Bernoulli distribution:
\begin{equation}
\pi_W(p|x^{loc},x^{img})=\prod_{n=1}^N (s^n)^{p^n}(1-s^n)^{1-p^n},
\end{equation}
where s is fusion result of the policy network, which can be formulated as
\begin{equation}
\begin{aligned}
s & =f(x^{loc},x^{img},W_{loc},W_{img})\\
& =sigmoid[\alpha W_{loc}x^{loc}+(1-\alpha)W_{img}x^{img}],
\end{aligned}
\end{equation}
where $f$ denotes the policy network with the weights $W_{loc}$ and $W_{img}$. $\alpha\in[0,1]$ is the fusion factor, $sigmoid(x)=\frac{1}{1+e^{-x}}$. $s=\{s^1,s^2,...,s^N\} \in R^{N}$ is the output of the policy network, and the $n$-th entry of $s$ ($s^n\in[0,1]$) indicates the probability that the $n$-th residual block is \emph{selected}. The policy $p\in\{0,1\}^N$ is obtained based on s, where 1 and 0 indicate \emph{selecting} and \emph{skipping} the corresponding block, respectively.

Based on the above,  we define the reward function to encourage more unique policy and minimal block usage, along with correct predictions. Our reward function is formulated as follows:
\begin{equation}
\label{E_reward}
R(p) = \left\{ \begin{array}{ll}
 \theta_s(1-(\frac{|p|_0}{N})^2) + \theta_d(1-(1-u)^2) & \textrm{if correct}\\
-\lambda & \textrm{otherwise},
\end{array} \right.
\end{equation}
where $\frac{|p|_0}{N}$ is the percentage of reserved blocks, implying the sparsity of the pre-trained ResNet, and $u$ reflects the difference from other $M-1$ policies, promoting diverse policies. When a prediction is correct, a large positive reward is offered to the policy network, while $\lambda$ is applied to penalize incorrect predictions. $\theta_s$ and $\theta_d$ represent the weights.

To learn the policy network, we maximize the following objective function \cite{wu2018blockdrop}:
\begin{equation}
\label{objective_function}
J=E_{p\sim\pi_W}[R(p)]
\end{equation}

\subsubsection{Sparsity of the Inference Path}\label{section3.2.1}
We imply that  the term $\theta_s(1-(\frac{|p|_0}{N})^2) $ in Eqn.~(\ref{E_reward}) increases the selection of the inference path,  so as to be diversified. It is also closely related to the complexity for the inference path selection. We deeply study that by further define $Pr$ that measure the sparsity of the selected inference path, as follows:
\begin{equation}
Pr= \frac{|p|_0}{N}
\end{equation}
We measure the proportion $Pr$, such as 15/21 for $|p|_0=15$ and $N=21$. $Pr$ also implies the computational complexity for each image during the inference, where a smaller value means lower complexity. For M$^2$Net, we adopt the average $Pr$ of the inference paths to measure the efficiency.

\begin{figure}[t]
\centering
\includegraphics[width=0.45\columnwidth]{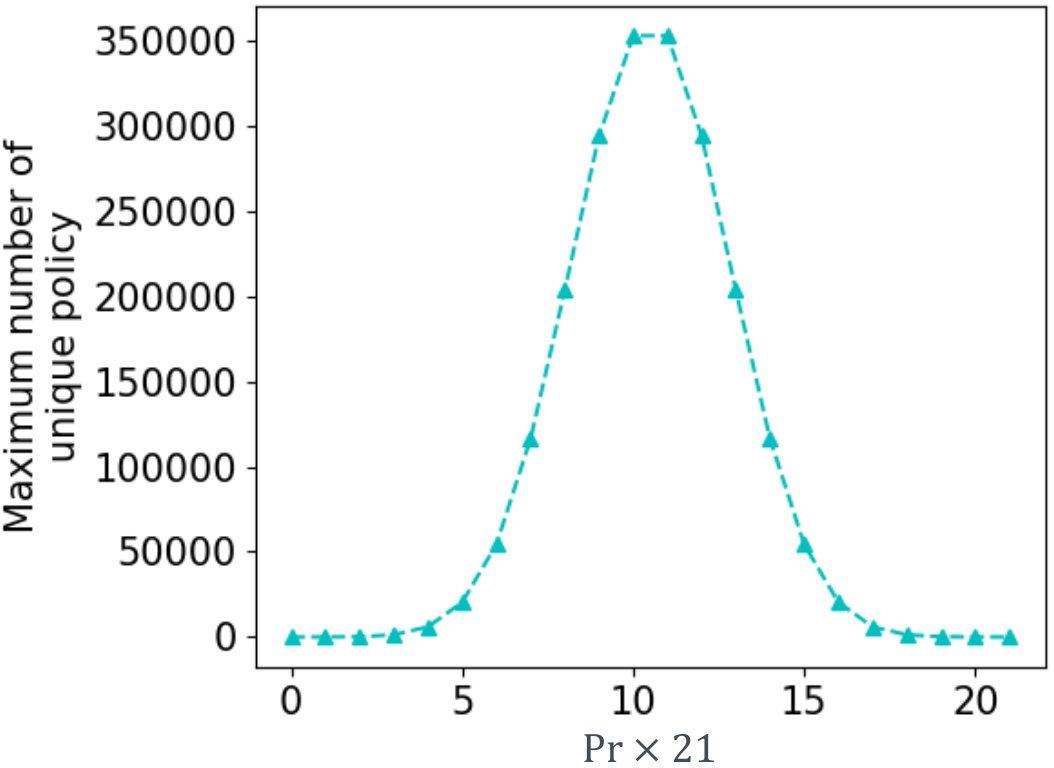}
\caption{The maximum number of unique policies under different $Pr$. When $Pr>10/21$, the upper bound of the available policy increases given $Pr$ decreases. Besides, the upper bound is much higher around $Pr=10/21$ than others.}
\label{number_unique_policy}
\end{figure}

We also obverse that the sparsity has non-negligible influence on the maximum number of unique policies, which can be computed by $C_{N}^{|p|_0}$,\emph{ e.g.,}  $C_{21}^{15}$. Fig.\ref{number_unique_policy} describes the relationship between them.  We observe that the maximum number of unique policies is of great difference under different $Pr$, and the peak value is obtained around $Pr=10/21$.

\subsection{Learning Paramters of M$^2$Net}\label{section3.2.2}
To learn two subnetwork weights  $W_{loc},W_{img}$ of the policy network, we compute the gradients of $J$ as motivated by \cite{wu2018blockdrop}. The gradients can be represented as:
\begin{equation}
\nabla_WJ=E[A\nabla_W\sum_{n=1}^{N}log[s^n p^n+(1-s^n)(1-p^n)]]
\end{equation}
where $A=R(p)-R(\tilde{p})$ and $\tilde{p}$ is the maximally probable configuration, i.e., $\tilde{p}^n=1$ if $s^n>0.5$, and $\tilde{p}^n=0$ otherwise \cite{rennie2017self}. $W$ denotes the weights of the policy network, which contains $W_{loc}$ and $W_{img}$.

We can further derive the gradients of $J$ with respect to $W_{loc}$ and $W_{img}$ as follows:
\begin{equation}
\nabla_{W_{loc}}J=x^{loc}\nabla_{W_{loc}x^{loc}+W_{img}x^{img}}J
\end{equation}

\begin{equation}
\nabla_{W_{img}}J=x^{img}\nabla_{W_{loc}x^{loc}+W_{img}x^{img}}J
\end{equation}

\subsection{More Insights among Reward Function, Policy Diversity and Recognition Accuracy}\label{More_Insights}
\begin{figure}[t]
\centering
\includegraphics[width=1.0\columnwidth]{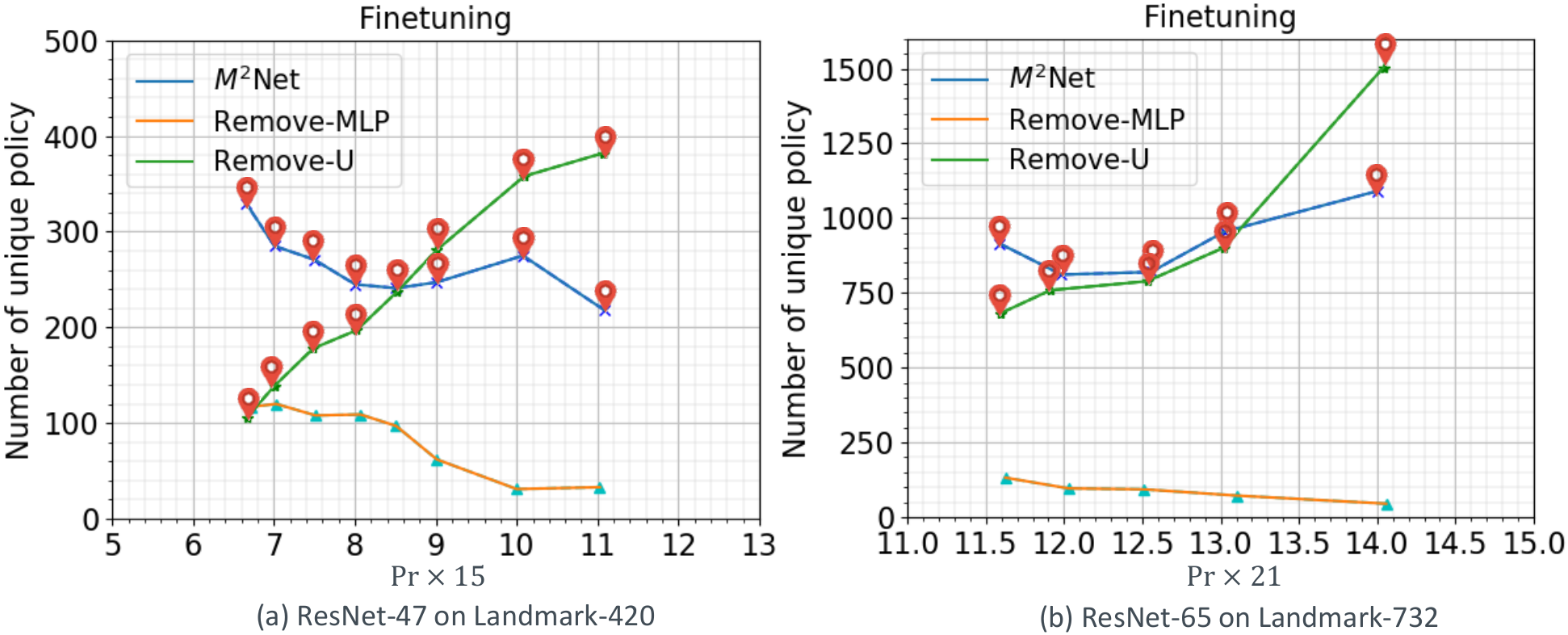}
\caption{Policy diversity. (a)(b) shows policy diversity of the policy network during finetuning process. M$^2$Net obtains more diverse policy compared to Remove-MLP and Remove-U as $Pr$ decreases.}
\label{Policy_diversity}
\end{figure}

As shown in Fig.\ref{fig2}, the reward function, policy diversity and recognition accuracy of M$^2$Net can inherently promote each other.  To better understand this, we discuss the relationship between each pair of them, respectively, which is shown below:

\begin{itemize}
\item \underline{Policy Diversity and Recognition Accuracy}:

For M$^2$Net, the multiple output policies of the policy network select the diverse inference path in the recognition network.  Besides, geographic information together with visual information are exploited to increase policy diversity. Combining Fig.\ref{Policy_diversity} with Fig.\ref{Test_Accuracy}(d) shows that M$^2$Net achieves larger policy diversity and higher recognition accuracy,  than the model that removed geographic information, implying that diverse policy can improve the recognition accuracy.
\item \underline{Recognition Accuracy and Reward Function}:

For Eqn.(\ref{E_reward}), when a prediction is correct, the reward function will generate a large reward value. The higher the recognition accuracy, the more correct predictions, resulting in more diverse reward values. That indicates that the larger recognition accuracy provides diverse reward values for the policy network.
\item \underline{Reward Function and Policy Diversity}:

To maximize Eqn.(\ref{objective_function}), the reward function as Eqn.(\ref{E_reward}) will encourage a smaller $Pr$ and diverse policy set.  We also find that the smaller the value of $Pr$, the greater the maximum number of unique policies, within a certain range, as shown in
Fig.\ref{number_unique_policy}. Thus the selection set of the output policy is expanded, leading to larger possibilities for diverse policies. This naturally indicates that the reward function $R$ can encourage larger policy diversity.
\end{itemize}

\section{Experiment}\label{section4}
In this section, we experimentally validate the effectiveness of M$^2$Net, including the comparisons with state-of-the-arts and comprehensive ablation studies.
\subsection{Experiment Setup}\label{section4.1}
\subsubsection{Datasets}\label{section4.1.1}
\begin{figure}[t]
\centering
\includegraphics[width=1.0\columnwidth]{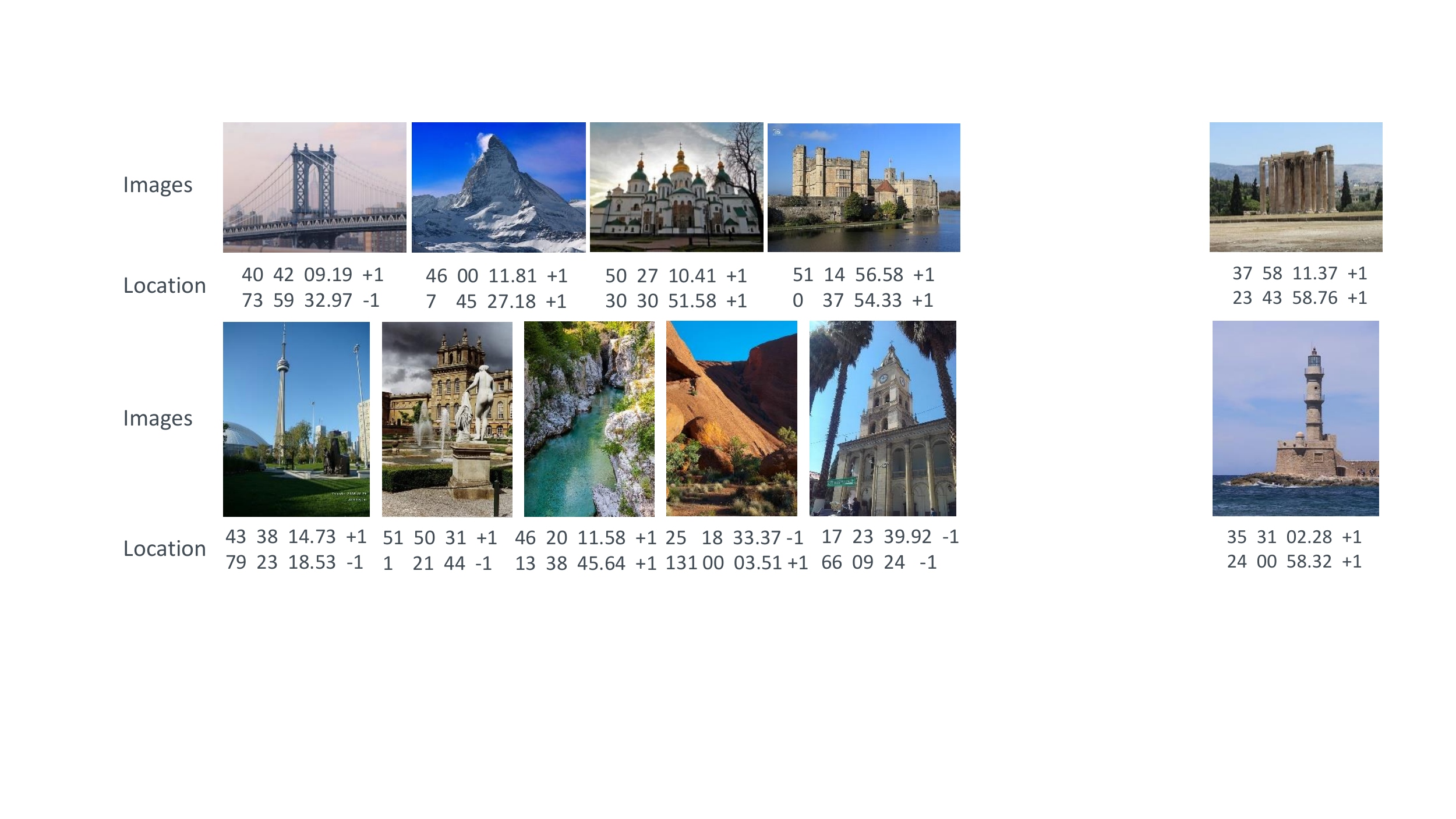}
\caption{Examples of landmark images with geographic information from our datasets, which contain artificial building and natural landscape.}
\label{Examples}
\end{figure}

We construct two landmark classification datasets: Landmark-420 and Landmark-732, where each image is picked from the Google-Landmark-v2 \cite{noh2017large}, which contains 5M images labeled for 200k unique landmarks ranging from artificial buildings to natural landscapes. We show the examples in Fig.\ref{Examples}. Each landmark image contains geographic information represented with latitude and longitude. The geographic location information is encoded as an 8-dimension vector, as stated in the Section ~\ref{geoinfo}.

In summary:
\begin{itemize}
\item The Landmark-420 dataset consists of 165,000 colored images, with 150,000 labeled landmark images  from 732 classes for training and 15,000 for test.
\item The Landmark-732 dataset consists of 380,000 labeled training landmark images across 732 categories and 40,000 images for test.
\end{itemize}

\begin{table}[ht]
\begin{center}
\begin{tabular}{|c|c|c|}%
    \hline
    Layer      & Input              & Output  \\  \hline \hline
    FC1        & 8                & 128 \\
    FC2        & 128            & 256 \\
    FC3        & 256            & 256 \\
    FC4        & 256            & 128 \\
    FC5        & 128            & K \\ \hline
\end{tabular}
\end{center}
\caption{ Multi-Layer Perceptron (MLP) architecture dealing with geographic information. K is set to 9, 15 and 21 for different pretrained ResNets, respectively.}
\label{MLP}
\end{table}

\subsubsection{Policy Network Architecture}\label{section4.1.3}
Our policy network combines geography locations with landmark images, with two subnetworks, as shown in Fig.\ref{Encoding}(b). For geographic locations, a 5-layer Multi-Layer Perceptron, as shown in Table \ref{MLP}, is adopted to extract the geographic location features. For landmark images, we utilize a ResNet variant named ResNet-10, as shown in Table \ref{ResNet-10}, to extract the visual features, while capturing the feature vectors via a fully connected layer \cite{wu2018blockdrop}. To exploit image features and corresponding geographic location, we fuse the output vectors of two subnetworks with a scaling factor $\alpha$ to balance them. In our experiments, we empirically set $\alpha$ as 0.7.

\begin{table}[ht]
\begin{center}
\renewcommand{\multirowsetup}{\centering}
\begin{tabular}{|c|c|c|c|}%
    \hline
    block             &Layer            & Filter                                            & Stride  \\ \hline  \hline
                         &Conv2d         & $7\times7\times3\times64$          & 2  \\ \hline
                         &MaxPool2d   & $3\times3$                                   &  2 \\ \hline  \hline
    Residual       &Conv2d          & $3\times3\times64\times64$             &  1\\
    block             &Conv2d         & $3\times3\times64\times64$             & 1 \\  \hline \hline
    \multirow{3}{1.2cm}{Residual block}   &
                                                       Conv2d           & $3\times3\times64\times128$           & 2 \\
                                                       &Conv2d         & $3\times3\times128\times128$         &  1\\
                                                       &Downsample & $1\times1\times64\times128$           & 2\\ \hline \hline
    \multirow{3}{1.2cm}{Residual block}   &
                                                       Conv2d            & $3\times3\times128\times256$          &  2\\
                                                       &Conv2d          & $3\times3\times256\times256$          &  1\\
                                                       &Downsample  & $1\times1\times128\times256$          & 2 \\ \hline \hline
    \multirow{3}{1.2cm}{Residual block}   &
                                                       Conv2d            & $3\times3\times256\times512$          & 2 \\
                                                       &Conv2d          & $3\times3\times512\times512$          & 1 \\
                                                       &Downsample  & $1\times1\times256\times512$          & 2\\ \hline \hline
                                                       &AvgPool2d     & $4\times4$                                          &  4\\ \hline
                                                      &Linear             & $512\times$ K                                   &-  \\ \hline
\end{tabular}
\end{center}
\caption{Residual network architecture extracting landmark image features. K is set to 9, 15 and 21 for three pretrained ResNets, respectively.}
\label{ResNet-10}
\end{table}

\subsubsection{Recognition Network}\label{section4.1.2}
We adopt three variants of ResNet \cite{he2016deep}, named ResNet-29, ResNet-47 and ResNet-65, respectively, which start with a convolutional layer followed by 9, 15 and 21 residual blocks that are organized into three blocks, respectively. Down-sampling layers are evenly inserted into them. To match our datasets, we make up a fully connected layer with 420/732 neurons at the end of the network. Each of the residual blocks adopts the bottleneck design \cite{he2016deep}, including the three convolutional layers. To obtain a pretrained ResNet, we train three ResNets from scratch on Landmark-420 and Landmark-732.

\begin{table}[h]
\footnotesize
\begin{center}
\begin{tabular}{|c|c|c|c|}%
    \hline
    \multirow{2}{*}{Method} &
    \multicolumn{2}{c|}{Landmark-420} &
    \multicolumn{1}{c|}{Landmark-732} \\
    \cline{2-4}
                    & ResNet-29 & ResNet-47 & ResNet-47   \\ \hline \hline
    Baseline        & $77.5\%$ & $78.8\%$ & $81.1\%$          \\  \hline \hline
    MobileNet \cite{howard2017mobilenets}       & $77.9\%$ & $79.9\%$ & $82.1\%$           \\  \hline \hline
    ShuffleNet(G2) \cite{zhang2018shufflenet} & $78.1\%$ & $79.1\%$ & $81.3\%$            \\
    ShuffleNet(G4) \cite{zhang2018shufflenet} & $75.8\%$ & $78.1\%$ & $80.0\%$           \\  \hline \hline
    HetConv(P2) \cite{Singh2019HetConv}   & $77.7\%$ & $79.1\%$ & $81.0\%$          \\
    HetConv(P4) \cite{Singh2019HetConv}   & $76.6\%$ & $76.7\%$ & $81.3\%$          \\  \hline \hline
    BlockDrop \cite{wu2018blockdrop}      & $77.6\%$ & $78.3\%$ & $81.5\%$           \\  \hline \hline
    M$^2$Net            & $\textbf{78.7\%}$ & $\textbf{83.6\%}$ & $\textbf{83.8\%}$    \\ \hline
\end{tabular}
\end{center}
\caption{Accuracy comparison with the state-of-the-arts.}
\label{Acc_comparison}
\end{table}

\subsection{Comparison with the State-of-the-Arts}
First of all, we compare the proposed M$^2$Net with several compressed network models:
\begin{itemize}
\item Baseline: it includes two variants of ResNet \cite{he2016deep}, named ResNet-29, ResNet-47.
\item MobileNet \cite{howard2017mobilenets}: it introduces a depth-wise separable convolution and $1\times1$ convolution to speed up convolutional computation.
\item ShuffleNet \cite{zhang2018shufflenet}: Based on MobileNet, it proposes a group convolution and shuffle operation to replace the $1\times1$ convolution.
\item HetConv \cite{Singh2019HetConv}: it proposes to replace $3\times3$ kernels with $1\times1$ kernels to speed up the convolution operation.
\item Blockdrop \cite{wu2018blockdrop}: it proposes a policy network based on reinforcement learning to output a set of binary decisions for dynamically dropping blocks in a pre-trained ResNet, where the policy network takes images as the inputs.
\end{itemize}

We implement the methods with the backbone network ResNet on Landmark-420 and Landmark-732. In particular, for ShuffleNet, G2 and G4 denote that the group number of $1\times1$ convolution is 2 and 4. While for HetConv, P2 and P4 indicate that the proportion of $3\times3$ kernels in a convolutional filter are 1/2 and 1/4.

\begin{figure}[t]
\centering
\includegraphics[width=1.0\columnwidth]{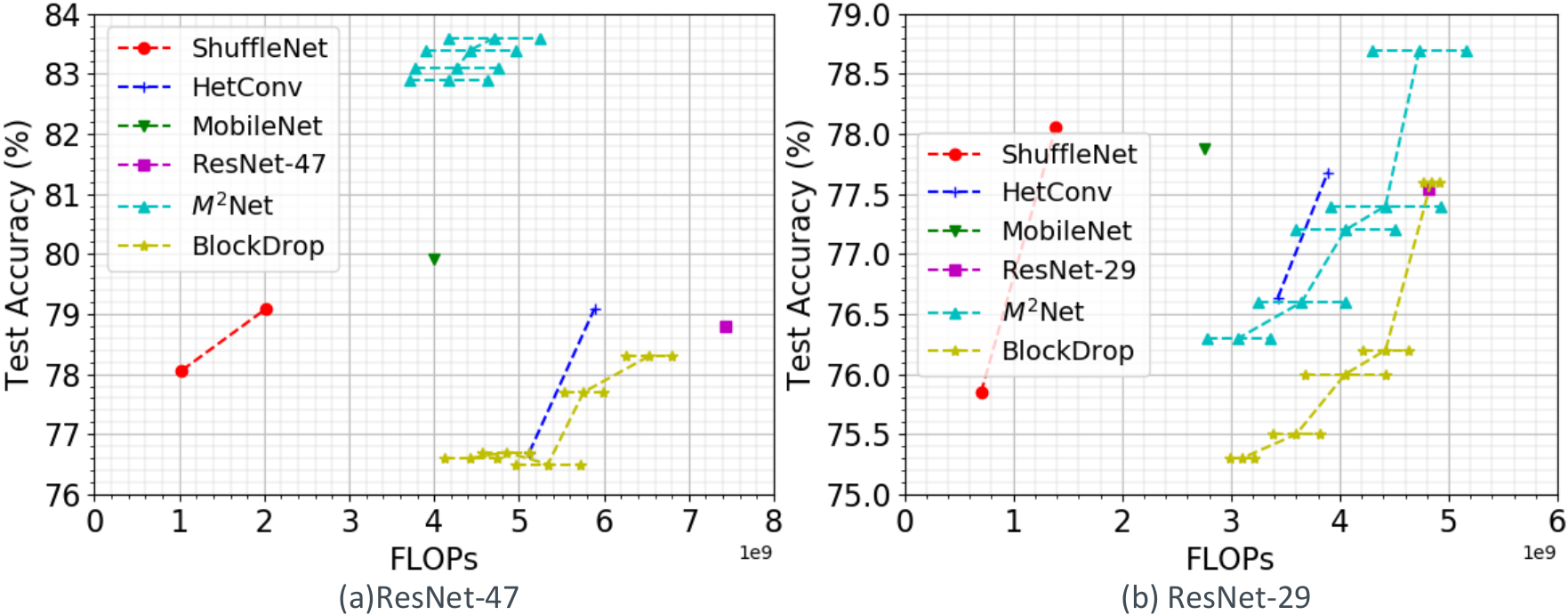}
\caption{Comparison of FLOPs with the state-of-the-arts on Landmark-420. Our M$^2$Net(cyan) outperforms other techniques with comparable computational complexity.}
\label{FLOPs}
\end{figure}

Table \ref{Acc_comparison} presents the results on Landmark-420 and Landmark-732. The results show that M$^2$Net outperforms other methods by large margins. We observe that our best model achieves $4.8\%$ accuracy gain over the baseline. When the recognition network changes from ResNet-29 to ResNet-47, our advantage is more obvious. This is because the higher number of blocks in ResNet-47 offer more possible options ($2^{15}$ $vs$ $2^9$) for the inference path. Thus, the policy network can generate more unique policies, and diverse inference paths can further be constructed.

Fig.\ref{FLOPs} presents the comparison of FLOPs with the state-of-the-art methods on Landmark-420. With ResNet-47, M$^2$Net obtains comparable computational complexity to MobileNet ($4.00 \times 10^9$ $vs$ $4.27 \times 10^9$) under the same level of accuracy. For some samples, M$^2$Net can even achieve faster inference speed than MobileNet. With ResNet-47, M$^2$Net achieves better accuracy than HetConv with the same level of computational efficiency. Compared to BlockDrop, M$^2$Net achieves higher computational efficiency with a larger variance, such as $4.42 \times 10^9 \pm 5.06 \times 10^8$ $vs$ $4.42 \times 10^9 \pm 2.05 \times 10^8$ with ResNet-29. Different from the static methods (MobileNet, ShuffleNet, HetConv), the computation efficiency of M$^2$Net is dynamically changeable for each input as shown in Fig.\ref{FLOPs}, which indicates that M$^2$Net can allocate computing resources reasonably.

\subsection{Ablation Study}
To better validate each module of M$^2$Net, we conduct a series of ablation studies in the next sections.
\subsubsection{With MLP vs Without MLP}
We explore the effect of geographic information in the policy network. We study the effect of the MLP. Without the MLP, the input of the policy network only contains images; we denote it as Remove-MLP for simplicity.

\begin{figure}[t]
\centering
\includegraphics[width=1.0\columnwidth]{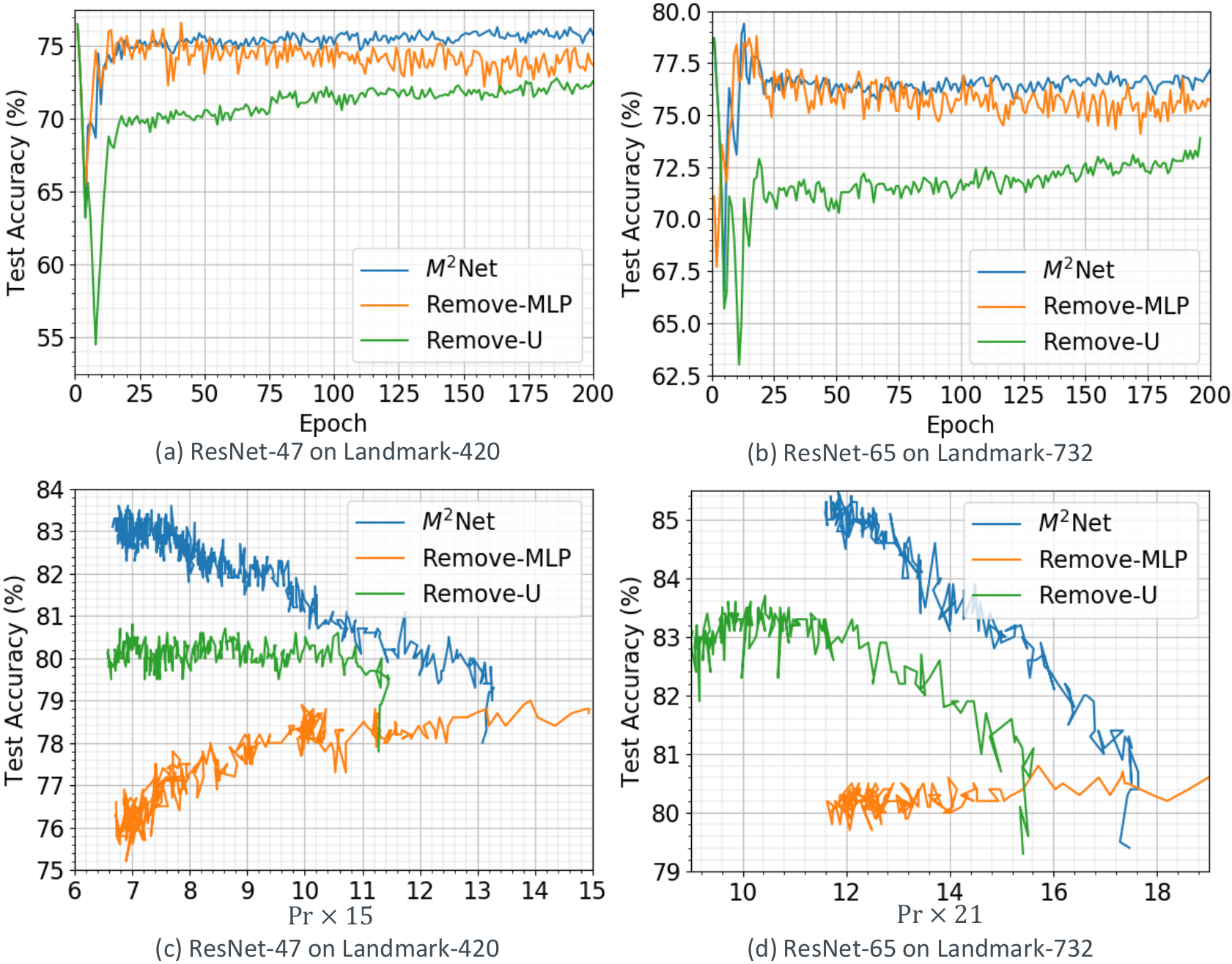}
\caption{Comparison of Test Accuracy. (a)(b) indicate the accuracy comparison for training the policy network. (c)(d) indicate the relationship between $Pr$ and test accuracy during finetuning. $Pr$ in (c)(d) represents the number of block retained on average. Smaller $Pr$ means less computational cost. We observe that M$^2$Net(blue) achieves obvious accuracy gains compared to Remove-MLP and Remove-U.}
\label{Test_Accuracy}
\end{figure}

The orange curves in Fig.\ref{Policy_diversity} and Fig.\ref{Test_Accuracy} present the results of Remove-MLP. From Fig.\ref{Policy_diversity}, we observe that the policy diversity of M$^2$Net is much higher than that of Remove-MLP($329$ $vs$ $117$ and $811$ $vs$ $96$), implying that combining geographic information with visual information can increase the number of unique policy. Besides, in Fig.\ref{Test_Accuracy}(c)(d), M$^2$Net achieves a large accuracy gain over Remove-MLP and keeps stable accuracy(83\% and 85\%). Thereby, we conclude that diverse policies contribute to improving the recognition accuracy.

In addition, Fig.\ref{Test_Accuracy}(c)(d) show that the accuracy of Remove-MLP sharply reduces as $Pr$ decreases, while that of M$^2$Net can be improved, which means that M$^2$Net can keep a good balance between accuracy and computational efficiency. In particular, Fig.\ref{Test_Accuracy}(c)(d) report that the accuracy curves reach a peak around $Pr=7/15$ or $11/21$, where the selection space of output policy  is largest as shown in Fig.\ref{number_unique_policy}.

\subsubsection{Reward Function}
We also explore the effect of uniqueness $U$ in the reward function. In the experiments, we remove the second term of Eqn.(\ref{E_reward}) and denote them as Remove-U for simplicity.

The green curves in Fig.\ref{Policy_diversity} and Fig.\ref{Test_Accuracy} present the results of Remove-U. Fig.\ref{Policy_diversity} shows that M$^2$Net can keep higher diversity compared to Remove-U when $Pr$ decreases, which indicates that the reward function with $U$ contributes to promoting the difference between output policies and increasing policy diversity. In addition, the accuracy of Remove-U decreases by a large margin($3\%$ and $2\%$) compared to M$^2$Net, as shown in Fig.\ref{Test_Accuracy}.

\begin{figure}[t]
\centering
\includegraphics[width=1.0\columnwidth]{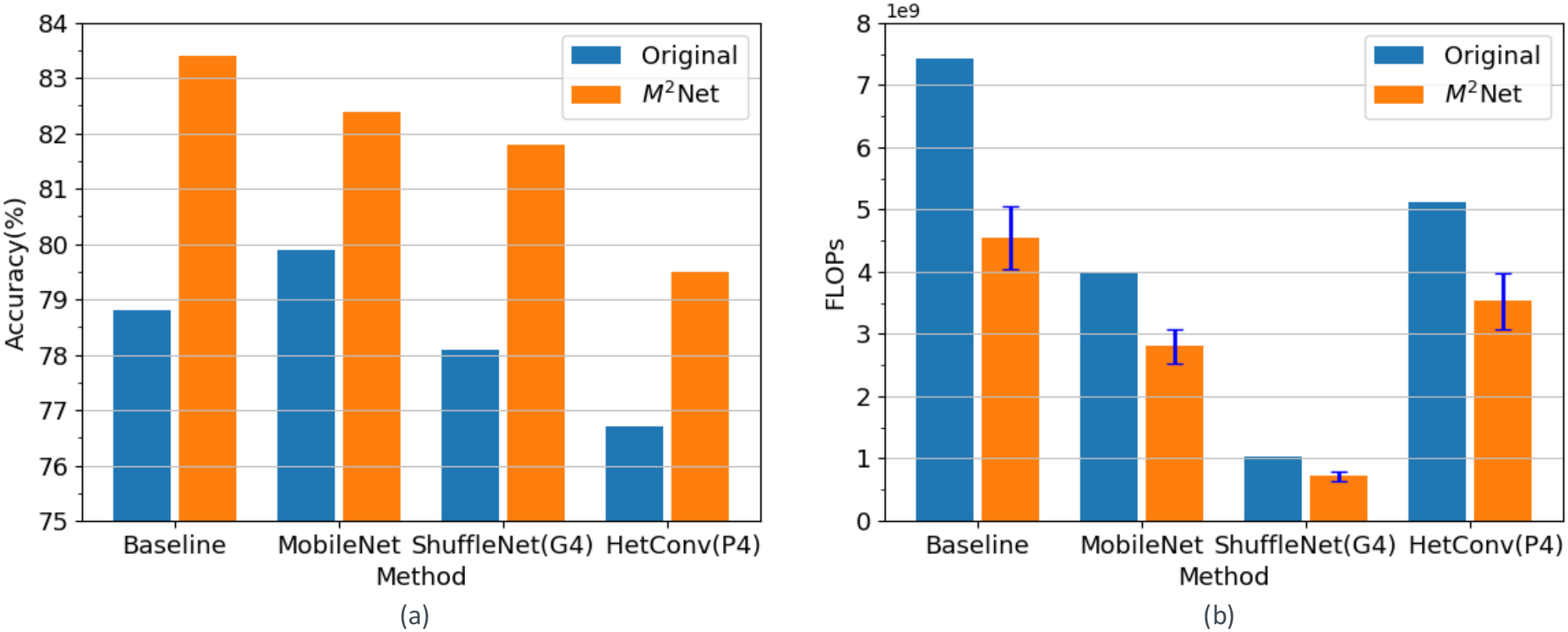}
\caption{Comparison of accuracy and FLOPs on Landmark-420. The orange denotes the results after adopting the backbone networks of other models as recognition network of M$^2$Net.}
\label{Compatibility}
\end{figure}

\subsection{Compatibility with State-of-the-Arts}
In the above experiments, we adopt the traditional pretrained ResNet as the recognition network. Actually, M$^2$Net is complementary to other methods, the backbone networks of which can also be used as our recognition network. We conduct the experiments on Landmark-420, where ResNet-47 is adopted as the baseline. Fig.\ref{Compatibility} presents the results of M$^2$Net after replacing the recognition network. We observe that not only is the recognition accuracy greatly improved, but the computational complexity is further reduced after applying our architecture to other methods. For MobileNet, ShuffleNet(G4) and HetConv(P4), the recognition accuracy is raised by $3.13\%$, $4.74\%$ and $3.65\%$, and the FLOPs is reduced by $30\%$, $30.3\%$ and $31.1\%$, respectively.

\section{Conclusion}
In this paper, we propose a novel moving-mobile net, named M$^2$Net,  for dynamically selecting  inference path for efficacy landmark recognition. Unlike existing methods, the geographic information of the landmarks is exploited to train a policy network, where the output served as the input to our proposed reward function, which can further promote the diverse selection of the inference path of M$^2$Net, so as to improve the performance. To validate the advantages of M$^2$Net, we create two landmark datasets with geo-information, over which extensive experiments are conducted.  The results validate the superiority of our method in terms of recognition accuracy and efficiency.

{\small
\bibliographystyle{ieee_fullname}
\bibliography{Reference}

\begin{thebibliography}{10}\itemsep=-1pt

\bibitem{cai2017deep}
Zhaowei Cai, Xiaodong He, Jian Sun, and Nuno Vasconcelos.
\newblock Deep learning with low precision by half-wave gaussian quantization.
\newblock In {\em Proceedings of the IEEE Conference on Computer Vision and
  Pattern Recognition}, pages 5918--5926, 2017.

\bibitem{chen2019layer}
Weijie Chen, Yuan Zhang, Di Xie, and Shiliang Pu.
\newblock A layer decomposition-recomposition framework for neuron pruning
  towards accurate lightweight networks.
\newblock In {\em AAAI}, volume~33, pages 3355--3362, 2019.

\bibitem{Cheng2016Convolutional}
Tai Cheng, Xiao Tong, Xiaogang Wang, and E Weinan.
\newblock Convolutional neural networks with low-rank regularization.
\newblock {\em Computer Science}, 2016.

\bibitem{han2015deep}
Song Han, Huizi Mao, and William~J Dally.
\newblock Deep compression: Compressing deep neural networks with pruning,
  trained quantization and huffman coding.
\newblock {\em arXiv preprint arXiv:1510.00149}, 2015.

\bibitem{han2015learning}
Song Han, Jeff Pool, John Tran, and William Dally.
\newblock Learning both weights and connections for efficient neural network.
\newblock In {\em Advances in neural information processing systems}, pages
  1135--1143, 2015.

\bibitem{he2016deep}
Kaiming He, Xiangyu Zhang, Shaoqing Ren, and Jian Sun.
\newblock Deep residual learning for image recognition.
\newblock In {\em CVPR}, pages 770--778, 2016.

\bibitem{He:2018:SFP:3304889.3304970}
Yang He, Guoliang Kang, Xuanyi Dong, Yanwei Fu, and Yi Yang.
\newblock Soft filter pruning for accelerating deep convolutional neural
  networks.
\newblock IJCAI'18, pages 2234--2240. AAAI Press, 2018.

\bibitem{he2019filter}
Yang He, Ping Liu, Ziwei Wang, Zhilan Hu, and Yi Yang.
\newblock Filter pruning via geometric median for deep convolutional neural
  networks acceleration.
\newblock In {\em Proceedings of the IEEE Conference on Computer Vision and
  Pattern Recognition}, pages 4340--4349, 2019.

\bibitem{hinton2015distilling}
Geoffrey Hinton, Oriol Vinyals, and Jeff Dean.
\newblock Distilling the knowledge in a neural network.
\newblock {\em arXiv preprint arXiv:1503.02531}, 2015.

\bibitem{howard2017mobilenets}
Andrew~G Howard, Menglong Zhu, Bo Chen, Dmitry Kalenichenko, Weijun Wang,
  Tobias Weyand, Marco Andreetto, and Hartwig Adam.
\newblock Mobilenets: Efficient convolutional neural networks for mobile vision
  applications.
\newblock {\em arXiv preprint arXiv:1704.04861}, 2017.

\bibitem{huang2018data}
Zehao Huang and Naiyan Wang.
\newblock Data-driven sparse structure selection for deep neural networks.
\newblock In {\em Proceedings of the European Conference on Computer Vision
  (ECCV)}, pages 304--320, 2018.

\bibitem{li2016pruning}
Hao Li, Asim Kadav, Igor Durdanovic, Hanan Samet, and Hans~Peter Graf.
\newblock Pruning filters for efficient convnets.
\newblock {\em arXiv preprint arXiv:1608.08710}, 2016.

\bibitem{liu2017learning}
Zhuang Liu, Jianguo Li, Zhiqiang Shen, Gao Huang, Shoumeng Yan, and Changshui
  Zhang.
\newblock Learning efficient convolutional networks through network slimming.
\newblock In {\em Proceedings of the IEEE International Conference on Computer
  Vision}, pages 2736--2744, 2017.

\bibitem{noh2017large}
Hyeonwoo Noh, Andre Araujo, Jack Sim, Tobias Weyand, and Bohyung Han.
\newblock Large-scale image retrieval with attentive deep local features.
\newblock In {\em Proceedings of the IEEE International Conference on Computer
  Vision}, pages 3456--3465, 2017.

\bibitem{peng2018extreme}
Bo Peng, Wenming Tan, Zheyang Li, Shun Zhang, Di Xie, and Shiliang Pu.
\newblock Extreme network compression via filter group approximation.
\newblock In {\em Proceedings of the European Conference on Computer Vision
  (ECCV)}, pages 300--316, 2018.

\bibitem{rastegari2016xnor}
Mohammad Rastegari, Vicente Ordonez, Joseph Redmon, and Ali Farhadi.
\newblock Xnor-net: Imagenet classification using binary convolutional neural
  networks.
\newblock In {\em European Conference on Computer Vision}, pages 525--542.
  Springer, 2016.

\bibitem{rennie2017self}
Steven~J Rennie, Etienne Marcheret, Youssef Mroueh, Jerret Ross, and Vaibhava
  Goel.
\newblock Self-critical sequence training for image captioning.
\newblock In {\em Proceedings of the IEEE Conference on Computer Vision and
  Pattern Recognition}, pages 7008--7024, 2017.

\bibitem{Singh2019HetConv}
Pravendra Singh, Vinay~Kumar Verma, Piyush Rai, and Vinay~P. Namboodiri.
\newblock Hetconv: Heterogeneous kernel-based convolutions for deep cnns.
\newblock In {\em CVPR}, 2019.

\bibitem{sutton2018reinforcement}
Richard~S Sutton and Andrew~G Barto.
\newblock {\em Reinforcement learning: An introduction}.
\newblock MIT press, 2018.

\bibitem{veit2016residual}
Andreas Veit, Michael~J Wilber, and Serge Belongie.
\newblock Residual networks behave like ensembles of relatively shallow
  networks.
\newblock In {\em Advances in neural information processing systems}, pages
  550--558, 2016.

\bibitem{wang2018two}
Peisong Wang, Qinghao Hu, Yifan Zhang, Chunjie Zhang, Yang Liu, and Jian Cheng.
\newblock Two-step quantization for low-bit neural networks.
\newblock In {\em Proceedings of the IEEE Conference on Computer Vision and
  Pattern Recognition}, pages 4376--4384, 2018.

\bibitem{wang2018skipnet}
Xin Wang, Fisher Yu, Zi-Yi Dou, Trevor Darrell, and Joseph~E Gonzalez.
\newblock Skipnet: Learning dynamic routing in convolutional networks.
\newblock In {\em Proceedings of the European Conference on Computer Vision
  (ECCV)}, pages 409--424, 2018.

\bibitem{wu2018blockdrop}
Zuxuan Wu, Tushar Nagarajan, Abhishek Kumar, Steven Rennie, Larry~S Davis,
  Kristen Grauman, and Rogerio Feris.
\newblock Blockdrop: Dynamic inference paths in residual networks.
\newblock In {\em Proceedings of the IEEE Conference on Computer Vision and
  Pattern Recognition}, pages 8817--8826, 2018.

\bibitem{xu2019main}
Yinghao Xu, Xin Dong, Yudian Li, and Hao Su.
\newblock A main/subsidiary network framework for simplifying binary neural
  networks.
\newblock In {\em Proceedings of the IEEE Conference on Computer Vision and
  Pattern Recognition}, pages 7154--7162, 2019.

\bibitem{yim2017gift}
Junho Yim, Donggyu Joo, Jihoon Bae, and Junmo Kim.
\newblock A gift from knowledge distillation: Fast optimization, network
  minimization and transfer learning.
\newblock In {\em Proceedings of the IEEE Conference on Computer Vision and
  Pattern Recognition}, pages 4133--4141, 2017.

\bibitem{zhang2018shufflenet}
Xiangyu Zhang, Xinyu Zhou, Mengxiao Lin, and Jian Sun.
\newblock Shufflenet: An extremely efficient convolutional neural network for
  mobile devices.
\newblock In {\em Proceedings of the IEEE Conference on Computer Vision and
  Pattern Recognition}, pages 6848--6856, 2018.

\end{thebibliography}
}

\end{document}